\documentclass{amia}
\usepackage{lipsum} 
\usepackage{amsmath}
\usepackage{algpseudocode}
\usepackage{algorithm}
\usepackage{graphicx}
\usepackage{adjustbox}
\usepackage{color}
\usepackage{enumitem}
\usepackage{makecell}
\usepackage{tabularx}
\usepackage{booktabs} 
\usepackage{array}    
\usepackage{caption}  
\usepackage{hyperref} 
\usepackage{xcolor}
\hypersetup{
    colorlinks,
    linkcolor={red!50!black},
    citecolor={blue!10!black},
    urlcolor={blue!50!black}
}
\usepackage{paralist}
\usepackage{CJKutf8}

\setlist{topsep=0pt, leftmargin=*}

\begin{document}

 \title{Leveraging Social Determinants of Health in Alzheimer's Research Using LLM-Augmented Literature Mining and Knowledge Graphs}

\author{
Tianqi Shang, MS$^{1,*}$, Shu Yang, PhD$^{1,*}$, Weiqing He, MS$^{1}$, Tianhua Zhai, PhD$^{1}$, \\
Dawei Li, MS$^{2}$, Bojian Hou, PhD$^{1}$, Tianlong Chen, PhD$^{3}$, \\ 
Jason H. Moore, PhD$^{4}$, Marylyn D. Ritchie, PhD$^{1}$, Li Shen, PhD$^{1,\dag}$}

\def\thefootnote{${*}$}\footnotetext{These authors contributed equally to this work. $^\dag$Correspondence to li.shen@pennmedicine.upenn.edu.}

\institutes{
    $^1$ Unversity of Pennsylvania, Philadelphia, PA, USA. 
    $^2$ Arizona State University, Tempe, AZ, USA. 
    $^3$ University of North Carolina at Chapel Hill, Chapel Hill, NC, USA. 
    $^4$ Cedars Sinai Medical Center, West Hollywood, CA, USA 
}

\maketitle

\section*{Abstract}
\textit{Growing evidence suggests that social determinants of health (SDoH), a set of nonmedical factors, affect individuals' risks of developing Alzheimer's disease (AD) and related dementias.
Nevertheless, the etiological mechanisms underlying such relationships remain largely unclear, mainly due to difficulties in collecting relevant information.
This study presents a novel, automated framework that leverages recent advancements of large language model (LLM) and natural language processing techniques to mine SDoH knowledge from extensive literature and integrate it with AD-related biological entities extracted from the general-purpose knowledge graph PrimeKG. 
Utilizing graph neural networks, we performed link prediction tasks to evaluate the resultant SDoH-augmented knowledge graph.
Our framework shows promise for enhancing knowledge discovery in AD and can be generalized to other SDoH-related research areas, offering a new tool for exploring the impact of social determinants on health outcomes.
Our code is available at: \url{https://github.com/hwq0726/SDoHenPKG}.
}

\section*{Introduction}

As the most common type of dementia among the aging population globally, Alzheimer's disease (AD) is a complex neurodegenerative disorder characterized by progressive cognitive decline and memory loss~\cite{WHO2024Dementia}. 
The prevalence of AD and related dementias is projected to impact more than 100 million individuals worldwide by 2050~\cite{2022e105}. 
It presents critical challenges for patients and significant burdens for the public health system~\cite{AD2024facts}. 
To date, the underlying etiology of AD is still largely unknown, primarily due to the heterogeneity of the disease~\cite{Ferreira436, MURRAY2011785, Bao2024Employing}. 
It is widely believed that there is not a single cause of AD but rather a number of different factors that affect individuals differently~\cite{CDC2024}. 
These factors include both medical and nonmedical ones, which can have a profound effect on a person’s risk for AD. 
Since currently limited pharmacological treatments exist for AD~\cite{AD2024facts}, many previous studies have been attempting to identify and address risk factors that are modifiable to prevent or delay the disease progression.  

Social Determinants of Health (SDoH), a set of nonmedical factors, have been shown to explain many aspects of the heterogeneity in cognitive, functional, biomarker, and interventional outcomes in AD and its related dementias (ADRD)~\cite{MAJOKA2021922, Joshi2024Social, CDC2024SDoHAD}. 
According to the World Health Organization~\cite{WHO2024SDOH}, SDoH refers to conditions in which people are born, grow, work, live, and age, and the wider set of forces and systems shaping the conditions of daily life. 
Knowledge of SDoH in AD/ADRD may help identify and address modifiable determinants, which are essential for improving health outcomes. 
For example, low education level has been shown to be associated with a greater risk of AD, as have social isolation and loneliness~\cite{maccora2020does, donovan2020social}.
Food insecurity is also reported to increase the likelihood of cognitive impairment~\cite{na2020food,wong2016food}, while low socioeconomic status has been found to be associated with accelerated cognitive decline~\cite{cadar2018individual, ajnakina2020interplay}, etc..
Although growing research demonstrates the impact of various SDoH on the risk of AD/ADRD, little is known about the underlying mechanisms linking SDoH and the fundamental biological processes involved in the disease development. 
One major obstacle here is the difficulty in collecting relevant data~\cite{Murray2020upstream, baum2009changes}. 
SDoH information is mainly scattered across scientific literature and the health system like medical notes in electronic health records (EHR). 
Manual collection is time-consuming and unsustainable given the sheer volume of such data.  

Automated methods utilizing natural language processing (NLP) have been developed to address the challenges and provide more efficient ways to extract SDoH information~\cite{patra2021extracting, ong2024artificial}. 
Previous studies have utilized a diverse range of NLP methods to identify and extract SDoH information from text: from initial, simple rule-based keyword searching~\cite{winden2018evaluation, Hollister2017developmen, Wu2023natural} and classic machine learning models~\cite{amrit2017identifying, feller2020detecting} to more recent deep neural networks~\cite{lybarger2021annotating, HAN2022103984} and the latest transformer-based large language models (LLMs)~\cite{guevara2024large,tan2024large}. 
The text sources in these studies are mostly unstructured free text from physician notes, nurse notes, etc. stored in the EHR systems.
However, besides the heterogeneous and distributed nature of these data, they often involve patient-related sensitive information, which imposes privacy challenges. 
Moreover, they were originally designed for non-research purposes and documented inconsistently, which may not adequately capture high-quality information on social factors~\cite{adler2015patients, bettencourt2020discovering}.
As a complementary source to EHR data, scientific literature such as published, peer-reviewed articles in PubMed can provide high-quality health concepts and also evidence connecting SDoH to other factors. 
Besides, knowledge graph (KG), as a tool to provide comprehensive knowledge representation for heterogeneous data from different sources, has demonstrated potential in this direction~\cite{park2021discovering, bettencourt2020exploring, gleize2021social, chandak2023building, li2024dalkdynamiccoaugmentationllms,li2024contextualization}. 
For instance, previous work~\cite{gleize2021social} has developed KGs to better store and present SDoH-related knowledge for COVID-19 during the pandemic. 
Payal et al. constructed a general knowledge graph called PrimeKG for precision medicine analyses by integrating heterogeneous biomedical entities like genes, diseases, drugs, and so on from 20 different resources, which is widely used in existing research~\cite{chandak2023building}. However, few works focus on the synergistic of scientific literature and KG together for SDoH mining.

In this work, we present a novel framework of LLM-augmented literature mining to extract SDoH knowledge in the context of AD and integrate it with existing biomedical knowledge through knowledge graph construction and integration. 
We first collect relevant articles from PubMed and perform named entity recognition (NER) on the articles to identify biomedical entities and SDoH entities, utilizing OpenAI’s latest pretrained LLM GPT-4o~\cite{GPT-4o} as well as classic NLP methods. 
For SDoH and biomedical entities, we then employ GPT's language abilities to extract their potential relations and filter for co-occurrences to ensure accuracy, resulting in triplets of (biomedical entity, relation, SDoH entity) which form an AD-related SDoH knowledge graph. 
Furthermore, we integrate our SDoH knowledge graph with a biomedical subgraph derived from the general-purpose PrimeKG to tailor for AD-SDoH study.
Finally, following previous work~\cite{pu2023graph, li2022graph, Huang2023.03.19.23287458}, we assess our new knowledge graph regarding the impact of SDoH on knowledge discovery in AD, by evaluating the performance of graph embedding-based link prediction tasks with graph neural networks~\cite{kipf2016semi}.

In summary, our main contributions can be listed as follows: 
\begin{compactenum}
    \item We develop an automated pipeline for extracting SDoH from extensive literature using pretrained large language model and advanced NLP techniques. Although this study focuses on AD, the pipeline is general and can be well applied to other domains.
    \item We integrate the extracted AD-related SDoH knowledge triples with the entities in PrimeKG to get an SDoH augmented knowledge graph with additional knowledge of social determinants for AD.
    \item As a result, we assess the utility of our new knowledge graph for link prediction tasks with graph convolutional networks (GCNs) and demonstrate its potential for identifying novel connections through the added SDoH triples, offering new insights for research and intervention for AD. 
\end{compactenum}

\section*{Methods}
\subsection*{Literature Corpus Collection}
A total of 12,773 articles related to Alzheimer's disease and SDoH were collected from PubMed (until July 31 2024). We used querying combinations of ‘Alzheimer's disease’ with each of the keywords (according to Table~\ref{tab:sdohtype}) ‘SDoH’, ‘community’, ‘economics’, ‘education’, ‘environment’, ‘healthcare’, ‘neighborhood’, and ‘social’ to search on PubMed and filtered for articles published within the last five years (2019–2024), preprints excluded. We conducted our analysis on the abstracts of these articles for efficiency consideration. 

\subsection*{SDoH Knowledge Graph Construction}

\begin{figure}[h]
    \centering
    \includegraphics[width=\textwidth]
    {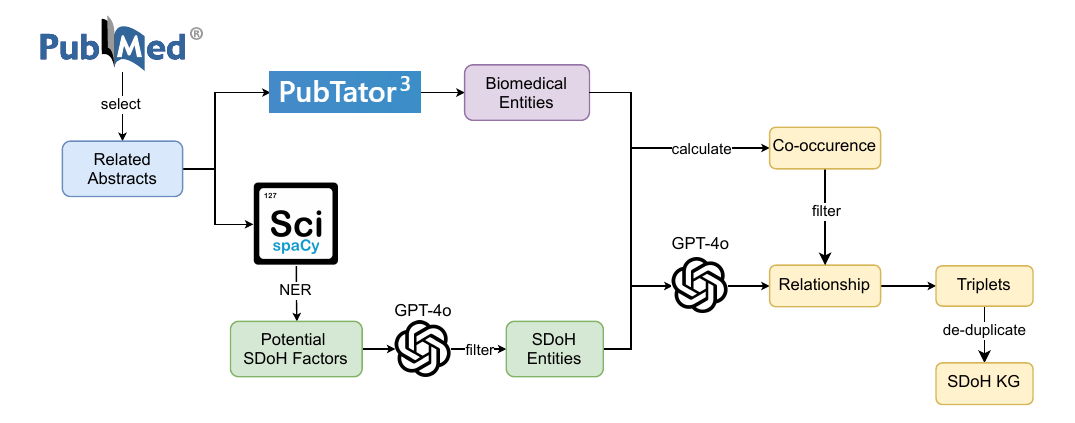}
    \caption{\small Overview of our framework for generating an AD-related SDoH knowledge graph from literature.}
    \label{fig:overview}
\end{figure}

Figure \ref{fig:overview} presents the overview of our framework for generating an AD-related SDoH knowledge graph from PubMed literature. This process begins with the selected abstracts as described in the above section, and each of the abstract is subsequently processed through the following multi-step procedures. \textbf{The detailed GPT-4o prompts with API calling, knowledge graph visualization, and all other codes are available on our \href{https://github.com/hwq0726/SDoHenPKG}{Github repo}, due to the space limit in the paper}.

\textbf{Entity Extraction:}  First, we use Pubtator 3.0~\cite{wei2024pubtator}, a biomedical text-mining tool, to extract bio-entities with predefined six types (i.e., genes, diseases, chemicals, mutations, species and cell lines) from each abstract. Meanwhile, we use scispaCy~\cite{neumann2019scispacy}, a robust NLP toolkit specialized for biomedical text processing, to perform named entity recognition (NER) to identify potential SDoH terms. At this stage, the identified entities are only considered potential SDoH candidates because the entity list may contain noise, including non-SDoH terms, incomplete phrases, or even biomedical entities already recognized by Pubtator 3.0. This necessitates further refinements. 

\textbf{Filtering and Classification:} To accurately identify the SDoH factors from the candidates, we leveraged OpenAI’s GPT-4o model~\cite{GPT-4o}. Here, this advanced language model was tasked with two primary functions: filtering and classification. Specifically, after removing all biomedical entities recognized by Pubtator 3.0, we prompted GPT-4o to classify the remaining potential SDoH candidates into one of the five primary types based on the SDoH framework provided by Healthy People 2030~\cite{healthypeople}, as detailed in Table~\ref{tab:sdohtype}. In the first filtering step, GPT-4o assessed each potential SDoH factor to determine whether it fits within any of the identified types. Entities that did not align with any type were discarded. For the remaining ones, GPT-4o further classified them into the most relevant subtype. Again, candidates that did not fit well into any subtype were removed. This dual-step filtering and classification process ensured that only the most confident SDoH factors were retained, each precisely categorized by type and subtype. \\

\begin{table}[ht]
\caption{\small Social Determinants of Health (SDoH) type and subtypes.}
\vspace{-0.3cm}
\resizebox{\textwidth}{!}{%
\begin{tabular}{lllll}
\toprule
\textbf{Economic Stability} &
  \textbf{\begin{tabular}[c]{@{}l@{}}Education Access \\ and Quality\end{tabular}} &
  \textbf{\begin{tabular}[c]{@{}l@{}}Health Care Access \\ and Quality\end{tabular}} &
  \textbf{\begin{tabular}[c]{@{}l@{}}Neighborhood and \\ Built Environment\end{tabular}} &
  \textbf{\begin{tabular}[c]{@{}l@{}}Social and \\ Community Context\end{tabular}} \\ \midrule 
Employment &
  \begin{tabular}[c]{@{}l@{}}Early Childhood Development \\ and Education\end{tabular} &
  Access to Health Services &
  \begin{tabular}[c]{@{}l@{}}Access to Foods That Support \\ Healthy Dietary Patterns\end{tabular} &
  Civic Participation \vspace{-0.3cm}\\ \\ \vspace{-0.3cm}
Food Insecurity     & Enrollment in Higher Education & Access to Primary Care & Crime and Violence       & Discrimination  \\ \\ \vspace{-0.3cm}
Housing Instability & High School Graduation         & Health Literacy        & Environmental Conditions & Incarceration   \\ \\ 
Poverty             & Language and Literacy          &       -                 & Quality of Housing       & Social Cohesion \\ \bottomrule
\end{tabular}%
}

\label{tab:sdohtype}
\end{table}

\textbf{Relation Extraction:} As we now have obtained the list of biomedical entities and the list of SDoH entities, for each biomedical and SDoH entity pair, we prompted GPT to go back to the abstract and analyze their potential relationships. The model was instructed to scrutinize the text and pinpoint explicit or implicit connections, such as causal relationships or associations. To further validate these relationships, we calculated the co-occurrence score for each entity pair. Formally, the co-occurrence score is defined as the ratio of the sentences in which both entities appeared together to the maximum frequency of either entity in the text. Only pairs with a co-occurrence score greater than 0.5 were retained, ensuring that the relationships we included in the knowledge graph were statistically significant and contextually relevant~\cite{theobald2009extraction, pu2023graph}. 

\textbf{Knowledge Graph Construction:} Finally, we integrate all the validated entity and relationship triplets derived from the abstracts to construct our SDoH knowledge graph. The graph consists of three types of triplets: 
\begin{compactitem}
    \item \textit{(biomedical entity, relation, SDoH entity):} Represents direct relationships between biomedical entities and SDoH factors. 
    \item \textit{(SDoH entity, subordinate, SDoH subtype):} Captures the hierarchical structure within SDoH factors, linking specific factors to their corresponding subtypes. 
    \item \textit{(SDoH subtype, subordinate, SDoH type):} Reflects the broader classification of SDoH subtypes into overarching types. 
\end{compactitem}

\begin{figure}[tb]
    \vspace{-0.1cm}
    \centering
    \includegraphics[width=0.85\textwidth]
    {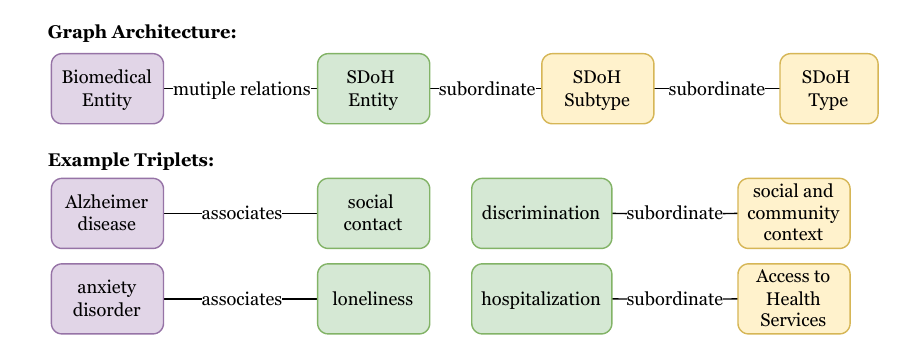}
    \caption{\small Architecture of the AD-related SDoH knowledge graph from literature mining, illustrated with example triplets.}
    \label{fig:architecture}
\end{figure}
Figure \ref{fig:architecture} illustrates the hierarchy architecture of the AD-related SDoH knowledge graph and some example triplets. After processing all the abstracts and de-duplicating the data, the resulting SDoH knowledge graph comprises a total of 4,058 edges and 1,364 unique SDoH entities/nodes. This graph serves as a comprehensive resource for understanding the complex interactions between social determinants and biomedical entities in the context of Alzheimer’s disease.  

\subsection*{Graph Evaluation}
To demonstrate the utility of our SDoH knowledge graph, we integrate our SDoH knowledge graph with the existing biomedical knowledge graph and introduce a series of link prediction experiments following previous work~\cite{pu2023graph, li2022graph, Huang2023.03.19.23287458} to prove that adding SDoH information can improve knowledge graph performances. 

\textbf{Integration with PrimeKG:} We integrated our SDoH knowledge graph with the widely-used, precision medicine-oriented knowledge graph, PrimeKG~\cite{chandak2023building} for demonstration. This integration will enable us to evaluate the impact of incorporating SDoH factors on knowledge discovery and prediction performance in AD research. Note that since our SDoH knowledge graph includes a unique identifier and source for every biomedical entity, it allows integration/merging with any other knowledge graphs using our method. For integration, we first identified and matched the biomedical entities in our SDoH knowledge graph with their corresponding entities in PrimeKG. Accurate entity mapping is crucial for ensuring consistency and maximizing the interoperability between the two graphs. As the two knowledge graphs share three types of overlapping entities: genes/proteins, drugs, and diseases, we collected data from multiple sources and created two mapping files to facilitate this matching process. 
\begin{compactitem}
    \item \textit{MeSH/OMIM ID to MONDO ID Mapping:} This file links disease entities identified by Medical Subject Headings (MeSH) or Online Mendelian Inheritance in Man (OMIM) IDs in SDoH KG to their corresponding entities in PrimeKG, which uses MONDO IDs. 
    \item \textit{MeSH ID to DrugBank ID Mapping:} This file links drug entities identified by MeSH IDs to their corresponding DrugBank IDs.  
\end{compactitem}

These mapping files allow us to seamlessly translate entity identifiers across different biomedical databases, ensuring accurate alignment. For entities in our SDoH graph that do not have a corresponding identifier mapping in PrimeKG, i.e. either lack inclusion in PrimeKG or do not possess the required type of identifier (e.g., a disease that has a MeSH ID but no MONDO ID because it is not cataloged in the MONDO database), we utilized the Unified Medical Language System (UMLS) database to retrieve the CUIs for the unmatched term and see whether they can be matched. For those nodes that still cannot be matched, we kept the original identifier from SDoH KG. This approach ensures that no valuable information is lost during the integration process. 

This ID mapping process significantly enhances the versatility of our SDoH knowledge graph. By enabling flexible switching between identifiers, our graph becomes more universal and can be easily stacked or merged with other medical knowledge graphs besides of PrimeKG. This interoperability makes our SDoH knowledge graph a powerful tool for broader biomedical research, allowing researchers to explore the impact of SDoH across various contexts and datasets. 

\textbf{Link Prediction Evaluation:} To show the efficacy of adding SDoH information to the knowledge graph, we present a series of link prediction experiments designed to demonstrate the utility of our SDoH KG. Link prediction in knowledge graphs is a critical task that aims to discover novel knowledge by inferring missing relationships or interactions between entities based on the observed data in the graph~\cite{pu2023graph, li2022graph, Huang2023.03.19.23287458}. When used for evaluation, the process typically involves a "mask and recover" procedure where certain relationships are deliberately masked (i.e., removed) from the training set of the graph, creating a partial graph. The task for the model is then to recover these masked edges based on the structural and semantic information it learns from the unmasked part of the graph. Models used for link prediction range from traditional statistical methods to advanced machine learning techniques, particularly graph convolutional networks (GCNs) \cite{kipf2016semi}. GCNs leverage node features and graph topology to learn embedded representations that facilitate predicting missing links/edges. 

Here for our task, formally, given a graph $\mathcal{G}=(\mathcal{V}, \mathcal{E})$ derived from our KG, which consists of a set of directed edge triples $e=(u, r, v) \in \mathcal{E}$, where $u, v \in \mathcal{V}$ are nodes representing SDoH or bio-entities and $r \in \mathcal{R}$ is a relation type. A GCN model learns to embed each node $v \in \mathcal{V}$ to a unit vector $\mathbf{z}_v$, and $\mathbf{z}_v$ can be used for downstream tasks like link prediction. Let $\mathcal{E}_{\text {train }} \subset \mathcal{E}$ denotes a set of observed training edges and let $\hat{\mathcal{E}}=\{\left( v_i, r, v_j\right): v_i, v_j \in\mathcal{V}, r \in R\} \backslash \mathcal{E}$ denote the set of negative edges that are not present in the true graph $\mathcal{G}$. Given $\mathcal{E}_{\text {train }}$ and a scoring function $s$, an ideal model should assign higher scores to true edges than to any negative (false) edges. Specifically, with cosine similarity, i.e. dot product here for unit vectors, as a typical option for $s$, we expect:
\begin{equation}
    \langle \mathbf{z}_v, \mathbf{z}_u \rangle > \langle \mathbf{z}_{v'}, \mathbf{z}_{u'} \rangle, \;\forall (u, r, v) \in \mathcal{E}, (u', r, v') \in \hat{\mathcal{E}}
\end{equation}
With this in mind, the model is trained to generate node representations that yield higher scores for positive (existing) edges and thus achieve the purpose of recovering the masked edges, which can then be used for discovering novel links. In our experiment, both the input and output node feature dimensions were set to 50. The training epoch was set to 200, and we repeated the experiment five times for each relation. The entire experiment was conducted on a Tesla P100 GPU, using Pytorch and the Deep Graph Library\footnote{https://www.dgl.ai/} for the GCN implementation. 

We conducted three types of link prediction tasks to assess the capabilities of our SDoH KG:  
\begin{compactitem}
    \item \textit{Random Mask Task:} In this task, we randomly masked some edges within the knowledge graph. The purpose of this experiment is to evaluate the general AD link prediction performance and overall robustness and accuracy with our SDoH-augmented KG.
    \item \textit{Targeted Genes Task:} This task specifically masked edges associated with the most related genes to AD.                                                        By focusing on these key genes, we aimed to assess how well the SDoH KG can help with recovering crucial biological relationships that are central to understanding AD. 
    \item \textit{Exploratory Prediction Task:} In this task, we generated and predicted edges that do not currently exist in the graph but could represent potential relationships worth exploring in future research. The exploratory nature of this task highlights the potential of the SDoH KG to aid in the discovery of new knowledge and insights within the AD domain. 
\end{compactitem}

\section*{Results}
\subsection*{Descriptive Statistics of The Knowledge Graphs}
Table \ref{tab:kgstat} provides a summary of the key statistics for our SDoH KG as well as our experiment knowledge graphs.  
\begin{compactitem}
    \item \textit{SDoH KG:}  This is the knowledge graph we constructed as described in the previous sections, focusing specifically on SDoH with AD related biomedical entities. 
    \item \textit{subPrimeKG:}  This is an AD-centered subgraph extracted from PrimeKG, which includes the two-hop neighbors of the AD node and is restricted to edges that contain at least one of "gene/protein," "disease," or "drug" type nodes.  
    \item \textit{SDoHenPK:}  This graph represents the integration of the SDoH KG with subPrimeKG as described in Integration with PrimeKG.  
\end{compactitem}

\begin{table}[ht]
\small
\caption{\small Statistics of the three knowledge graphs used in our study.}
\vspace{-0.3cm}
\centering
\begin{tabularx}{0.7\textwidth}{XXXX}
\toprule
\textbf{Graph} & \textbf{SDoH KG} & \textbf{subPrimeKG} & \textbf{SDoHenPKG} \\ 
\midrule
node & 2,109 & 17,368 & 19,190\\ 

edge & 4,058 & 245,252 & 249,267 \\ 
\bottomrule
\end{tabularx}
\label{tab:kgstat}
\end{table}
\subsection*{General AD Link Predictions with SDoH Augmentations}
As described earlier, to evaluate the general performance, we employed a "mask and recover" strategy and trained a GCN model for the link prediction task. Specifically, we compared the performance on subPrimeKG and SDoHenPK (Table \ref{tab:kgstat}). By comparing these two graphs, we aimed to determine whether the addition of SDoH knowledge improves link prediction performance. In this experiment, we focused on seven target relations (listed in Table 3) and, for each relation, randomly masked 20\% of the edges in both subPrimeKG and SDoHenPKG to create the training set, with the masked edges serving as the test set. We then trained a GCN model separately on subPrimeKG and SDoHenPKG, using different convolutional layers to extract feature information from various relations and aggregate it to obtain the final node features. After training, we evaluated and compared the models' performance on the test set. We adopted the standard Mean Reciprocal Rank (MRR) as our metric for comparison, a widely-used metric in knowledge graph evaluation~\cite{bordes2013translating,trouillon2016complex,sun2019rotate}. Specifically, we corrupted each test edge by replacing its tail with 20 randomly-sampled negative entities. The goal was to rank the true tail entities higher than the negative entities. Since SDoHenPKG is a featureless heterogeneous graph, we randomly generated vectors as the input node features. 

As shown in Table \ref{tab:result1}, by merging SDoH information into subPrimeKG, SDoHenPKG shows a significant improvement in the link prediction task across all selected relations compared to subPrimeKG. This enhancement is evident from the consistently higher MRR values observed in SDoHenPKG, along with the p-values that indicate these improvements are statistically significant for most relations. Significant improvements are observed in the Disease-Effect and Disease-Gene/Protein relations, indicating that SDoH factors may provide additional context for predicting disease progression and interactions. Similarly, improvements in Drug-Effect and Drug-Gene/Protein relations suggest that SDoH factors may impact drug efficacy, aligning with existing observations that SDoH can affect adherence, access, tolerance, and the overall clinical impact of dementia medications. 
While other relations show minor or insignificant improvements, the consistent pattern across key biomedical interactions highlights the relevance of SDoH in enhancing knowledge graph predictions. This result suggests that the integration of SDoH data appears to provide additional context that helps the model better infer missing links related to AD. \\

\begin{table}[ht] 
\caption{\small Results of the general AD link predictions task across seven relations. 
The MRR values are the averages of five repeated experiments with standard deviations. * indicates significance levels from a two-tailed paired t-test (p-value $<$ 0.05).}
\vspace{-0.3cm}
\centering

\resizebox{\textwidth}{!}{%
\small 
\begin{tabular}{lp{0.1\textwidth}p{0.1\textwidth}p{0.1\textwidth}p{0.1\textwidth}p{0.08\textwidth}p{0.155\textwidth}p{0.1\textwidth}}
\toprule

\textbf{} & \textbf{Anatomy-Gene/Protein} & \textbf{Disease-Drug} & \textbf{Disease-Effect} & \textbf{Disease-Gene/Protein} & \textbf{Drug-Effect} & \textbf{Biological Process-Gene/Protein} & \textbf{Drug-Gene/Protein} \\ \midrule
\begin{tabular}[c]{@{}c@{}}subPrimeKG \\ (MRR)\end{tabular} &
  \begin{tabular}[c]{@{}c@{}}0.923 ± \\ 0.004\end{tabular} &
  \begin{tabular}[c]{@{}c@{}}0.949± \\ 0.006\end{tabular} &
  \begin{tabular}[c]{@{}c@{}}0.789± \\ 0.021\end{tabular} &
  \begin{tabular}[c]{@{}c@{}}0.941± \\ 0.005\end{tabular} &
  \begin{tabular}[c]{@{}c@{}}0.56± \\ 0.009\end{tabular} &
  \begin{tabular}[c]{@{}c@{}}0.529± \\ 0.014\end{tabular} &
  \begin{tabular}[c]{@{}c@{}}0.636± \\ 0.031\end{tabular} \\
\begin{tabular}[c]{@{}c@{}}SDoHenPK \\ (MRR)\end{tabular} &
  \textbf{\begin{tabular}[c]{@{}c@{}}0.927± \\ 0.001\end{tabular}} &
  \textbf{\begin{tabular}[c]{@{}c@{}}0.958± \\ 0.003\end{tabular}} &
  \textbf{\begin{tabular}[c]{@{}c@{}}0.815± \\ 0.007\end{tabular}} &
  \textbf{\begin{tabular}[c]{@{}c@{}}0.946± \\ 0.005\end{tabular}} &
  \textbf{\begin{tabular}[c]{@{}c@{}}0.568± \\ 0.009\end{tabular}} &
  \textbf{\begin{tabular}[c]{@{}c@{}}0.654± \\ 0.136\end{tabular}} &
  \textbf{\begin{tabular}[c]{@{}c@{}}0.654± \\ 0.027\end{tabular}} \\
p-value        & 0.059      & 0.063      & 0.037 *    & 0.025 *    & 0.042 *    & 0.077      & 0.029 *    \\ \bottomrule
\end{tabular}%
}

\label{tab:result1}
\end{table}

\subsection*{Targeted Analysis of Key AD Genes}
Here, we consider a list of predefined AD-related genes and assess how well our SDoH KG can help with recovering targeted AD-related biological relationships. 
We selected 17 genes (listed in Table 4) from the top 20 AD-related genes, as three were absent in the PrimeKG~\cite{chandak2023building}. 
For each of them, we used STRING v12.0~\cite{szklarczyk2023string} to find its connection with other genes, focusing exclusively on experimentally validated edges. 
These edges were then used to create the masked datasets.
We follow the same experiment settings as the random mask task above, and also compare the performance of subPrimeKG and SDoHenPKG. Table \ref{tab:result2} presents the results of this task.

\begin{table}[tb]
\vspace{0.5cm}
\small
\caption{\small Link prediction results of the AD-related genes.}
\vspace{-0.3cm}
\centering
\resizebox{0.9\textwidth}{!}{%
\begin{tabular}{ccccccccc}
\toprule
\textbf{}    & \textbf{APOE}   & \textbf{TREM2}   & \textbf{CD33}  & \textbf{CLU}   & \textbf{BIN1}   & \textbf{CR1}   & \textbf{SORL1} & \textbf{CD2AP} \\ \midrule
subPrimeKG (MRR) & 0.527           & 0.717            & 0.457          & 0.604          & 0.524           & 0.133          & 0.551          & 0.375          \\
SDoHenPK (MRR)   & \textbf{0.547}  & \textbf{1.000}   & \textbf{0.528} & \textbf{0.673} & \textbf{0.603}  & \textbf{0.193} & \textbf{0.584} & \textbf{0.414} \\
p-value          & 0.038 *         & 0.013 *          & 0.052          & 0.070          & 0.044 *         & 0.049 *        & 0.069          & 0.063          \\ \bottomrule
(continued)      &                 &                  &                &                &                 &                &                &                \\ \toprule
\textbf{EPHA1}   & \textbf{INPP5D} & \textbf{SHARPIN} & \textbf{ABI3}  & \textbf{PLCG2} & \textbf{PICALM} & \textbf{C7}    & \textbf{SPI1}  & \textbf{PTK2B} \\ \midrule
0.438            & 0.578           & 0.670            & 0.177          & 0.674          & 0.271           & 0.506          & 0.569          & 0.574          \\
\textbf{0.571}   & \textbf{0.649}  & \textbf{0.750}   & \textbf{0.222} & \textbf{0.752} & \textbf{0.331}  & \textbf{0.638} & \textbf{0.693} & \textbf{0.656} \\
0.054            & 0.025 *         & 0.044 *          & 0.020 *        & 0.064          & 0.027 *         & 0.015 *        & 0.038 *        & 0.036 *        \\ \bottomrule
\end{tabular}%
}

\label{tab:result2}
\end{table}

As we can see in Table \ref{tab:result2}, SDoHenPKG, which incorporates SDoH, generally outperformed subPrimeKG in predicting gene relationships, as demonstrated by higher MRR values across all the selected genes. The p-values indicate statistically significant improvements in predictive performance for several genes with SDoHenPKG. Specifically, genes like TREM2, BIN1, CR1, SPI1, INPP5D, ABI3, and C7 showed significant improvements, validating the integration of SDoH into the KG. This result shows that SDoH knowledge augments gene-gene link predictions, suggesting that social factors influence gene interactions relevant to AD pathology, which strongly supports the benefit of incorporating SDoH data into etiology research, particularly for complex diseases like Alzheimer's where biological and social factors are closely interlinked.

\subsection*{Exploratory Analysis to Discover Novel Relationships}

Given the expensive costs and tremendous efforts required to study the relationships between biomedical entities in experiments, graph analysis represents a promising approach for inferring new relationships and discovering novel knowledge~\cite{tran2020heterogeneous,rao2018phenotype,valentini2014extensive,renaux2023knowledge}. To further explore the potential applications of our knowledge graph, we utilized it to predict potential interactions between previously unconnected nodes. In this task, we use the full graph as the training set, aiming to calculate the probability of each non-existent gene-SDoH and gene-gene edge. After training the GCN model as in previous tasks, we applied a sigmoid function to the GCN outputs to transform them into probabilities and obtain the likelihood of edge existence.

For the predicted gene-SDoH edges, we identified 36 AD-related genes that might be influenced by SDoH. SDoH affects these genes by potentially influencing their expression through epigenetic modifications driven by factors such as chronic stress, poor nutrition, and environmental exposures~\cite{notterman2015epigenetics}. The epigenetic changes might exacerbate gene dysfunction, leading to increased oxidative stress, inflammation, and metabolic disturbances that contribute to the development and progression of AD~\cite{adkins2023structural}. Among these genes, dysfunction of the LYZ gene, resulting in altered lysozyme levels, may contribute to AD etiology by affecting amyloid-beta aggregation and promoting neuroinflammation~\cite{udayar2022lysosomal}. SDoH might impact the LYZ gene by influencing gene expression and immune function through factors like poor nutrition, environmental exposures, and lifestyle choices such as smoking and alcohol use. These SDOH-induced changes might lead to altered lysozyme levels and immune dysregulation, contributing to chronic inflammation that increases the risk of AD. Addressing corresponding SDoH is likely essential to mitigate these effects and reduce the risk of AD. 

For the top 1,000 (0.0001\textperthousand) predicted gene-gene edges with the highest probabilities, we conducted a PubMed search to check each gene pair's co-occurrence and recorded the number of relevant articles. Notably, only 41 out of the 1,000 gene pairs had zero co-occurrence in the PubMed database, indicating that our predictions are both accurate and reasonable. The overlap of predicted gene-gene edges with existing literature validates the reliability of our SDoH-augmented knowledge graph, demonstrating that the inclusion of SDoH information contributes to predicting biologically meaningful relationships rather than introducing arbitrary or random edges. Most gene pairs exhibited relatively low frequency, suggesting that they have not been extensively studied, thereby highlighting opportunities for future research. Interestingly, one gene pair, IL1B and TNF, had the highest co-occurrence frequency of 215,719 but was not included in PrimeKG. This finding illustrates that our prediction task can also be used to identify omissions or gaps in existing knowledge graphs. The complete list of all the predicted edges is available at our \href{https://github.com/hwq0726/SDoHenPKG}{Github repo}.

\section*{Discussion and Conclusion}

In this study, we developed an automated pipeline for extracting SDoH from extensive AD-related literature by utilizing LLM and advanced natural language processing techniques. This approach enabled the identification, classification, and validation of SDoH factors from a vast body of literature. We also systematically organize relationships between SDoH factors and biomedical entities to build an SDoH knowledge graph. Additionally, we present the method for integration between our SDoH knowledge graph with established biomedical knowledge graphs to create a more comprehensive, SDoH-augmented knowledge graph. This enriched graph enhanced our understanding of the intersection between social determinants and biological factors in AD. Through the integration of these SDoH triples, we demonstrated the utility of the knowledge graph in link prediction tasks using graph neural networks, which not only improved the prediction of missing links but also uncovered novel connections that had not been previously explored. This approach offers valuable new insights for AD research, potentially informing future interventions and expanding the scope of disease understanding by emphasizing the role of social factors.

While our framework shows promise, several challenges remain. First, some SDoH factors were difficult to classify accurately, especially those that are ambiguous or context-dependent, which may affect the completeness and precision of the resulting knowledge graph. Additionally, the computational resources required for graph construction were substantial, due to the need to process duplicated content across large volumes of literature. This could limit the accessibility of the approach to researchers with constrained computational capabilities. Lastly, while our link prediction experiments demonstrated overall improvement, not all predictions reached statistically significant levels, likely due to the complexity of the relationships being modeled or limitations in the data itself. Future work could address these issues by refining the classification models, leveraging more robust computational resources, and incorporating additional data sources to enhance the predictive power and statistical significance of the results.

Our study opens several avenues for future research. One promising direction is the further exploration of the novel SDoH-gene and gene-gene connections identified in our exploratory task. These predictions could provide potentially critical insights into the biological mechanisms underlying AD for further experimental validations. Additionally, the framework we developed is generalizable to other diseases or health conditions where SDoH factors play a significant role, offering a new tool for investigating the complex interplay between social determinants and health outcomes. In conclusion, our work demonstrates the value of integrating SDoH into biomedical knowledge graphs and provides a novel approach to enhancing knowledge discovery in AD research. 
This research contributes a significant resource and methodology to the field, with broad implications for future studies in AD and beyond.

\subparagraph{Acknowledgments} This work was supported in part by the NIH grants P30 AG073105, U01 AG066833, U01 AG068057, and R01 AG071470. 

\makeatletter
\renewcommand{\@biblabel}[1]{\hfill #1.}
\makeatother

\small
\bibliographystyle{vancouver}
\bibliography{amia}  

\begin{thebibliography}{10}

\bibitem{WHO2024Dementia}
WHO. Dementia;.
\newblock Accessed: 2024-08-30.
\newblock \url{https://www.who.int/news-room/fact-sheets/detail/dementia}.

\bibitem{2022e105}
Nichols E, Steinmetz JD, Vollset SE, Fukutaki K, Chalek J, Abd-Allah F, et~al.
\newblock Estimation of the global prevalence of dementia in 2019 and forecasted prevalence in 2050: an analysis for the Global Burden of Disease Study 2019.
\newblock The Lancet Public Health. 2022;7(2):e105-25.

\bibitem{AD2024facts}
2024 Alzheimer's disease facts and figures.
\newblock Alzheimer's \& Dementia. 2024;20(5):3708-821.
\newblock Available from: \url{https://alz-journals.onlinelibrary.wiley.com/doi/abs/10.1002/alz.13809}.

\bibitem{Ferreira436}
Ferreira D, Nordberg A, Westman E.
\newblock Biological subtypes of Alzheimer disease.
\newblock Neurology. 2020;94(10):436-48.

\bibitem{MURRAY2011785}
Murray ME, Graff-Radford NR, Ross OA, Petersen RC, Duara R, Dickson DW.
\newblock Neuropathologically defined subtypes of Alzheimer's disease with distinct clinical characteristics: a retrospective study.
\newblock The Lancet Neurology. 2011;10(9):785-96.

\bibitem{Bao2024Employing}
Bao J, Lee B, Wen J, Kim M, Mu S, Yang S, et~al.
\newblock Employing Informatics Strategies in Alzheimer's Disease Research: A Review from Genetics, Multiomics, and Biomarkers to Clinical Outcomes.
\newblock Annual review of biomedical data science. 2024 06;7.

\bibitem{CDC2024}
CDC. Alzheimer's Disease and Healthy Aging: About Alzheimer’s Disease;.
\newblock Accessed: 2024-08-30.
\newblock \url{https://www.cdc.gov/aging/alzheimers-disease-dementia/about-alzheimers.html}.

\bibitem{MAJOKA2021922}
Majoka MA, Schimming C.
\newblock Effect of Social Determinants of Health on Cognition and Risk of Alzheimer Disease and Related Dementias.
\newblock Clinical Therapeutics. 2021;43(6):922-9.

\bibitem{Joshi2024Social}
Joshi P, Tampi R.
\newblock Social Determinants of Health for Alzheimer's Disease and Other Dementias.
\newblock Psychiatric Annals. 2024;54(7):e216-22.

\bibitem{CDC2024SDoHAD}
CDC. Social Determinants of Health and Alzheimer’s Disease and Related Dementias;.
\newblock Accessed: 2024-08-30.
\newblock \url{https://www.cdc.gov/aging/disparities/social-determinants-alzheimers.html}.

\bibitem{WHO2024SDOH}
WHO. Social determinants of health;.
\newblock Accessed: 2024-08-30.
\newblock \url{https://www.who.int/health-topics/social-determinants-of-health}.

\bibitem{maccora2020does}
Maccora J, Peters R, Anstey KJ.
\newblock What does (low) education mean in terms of dementia risk? A systematic review and meta-analysis highlighting inconsistency in measuring and operationalising education.
\newblock SSM-population health. 2020;12:100654.

\bibitem{donovan2020social}
Donovan NJ, Blazer D.
\newblock Social isolation and loneliness in older adults: review and commentary of a national academies report.
\newblock The American Journal of Geriatric Psychiatry. 2020;28(12):1233-44.

\bibitem{na2020food}
Na M, Dou N, Ji N, Xie D, Huang J, Tucker KL, et~al.
\newblock Food insecurity and cognitive function in middle to older adulthood: a systematic review.
\newblock Advances in Nutrition. 2020;11(3):667-76.

\bibitem{wong2016food}
Wong JC, Scott T, Wilde P, Li YG, Tucker KL, Gao X.
\newblock Food insecurity is associated with subsequent cognitive decline in the Boston Puerto Rican Health Study.
\newblock The Journal of nutrition. 2016;146(9):1740-5.

\bibitem{cadar2018individual}
Cadar D, Lassale C, Davies H, Llewellyn DJ, Batty GD, Steptoe A.
\newblock Individual and area-based socioeconomic factors associated with dementia incidence in England: evidence from a 12-year follow-up in the English longitudinal study of ageing.
\newblock JAMA psychiatry. 2018;75(7):723-32.

\bibitem{ajnakina2020interplay}
Ajnakina O, Cadar D, Steptoe A.
\newblock Interplay between socioeconomic markers and polygenic predisposition on timing of dementia diagnosis.
\newblock Journal of the American Geriatrics Society. 2020;68(7):1529-36.

\bibitem{Murray2020upstream}
Murray GF, Rodriguez HP, Lewis VA.
\newblock Upstream With A Small Paddle: How ACOs Are Working Against The Current To Meet Patients’ Social Needs.
\newblock Health Affairs. 2020;39(2):199-206.
\newblock PMID: 32011930.

\bibitem{baum2009changes}
Baum FE, B{\'e}gin M, Houweling TA, Taylor S.
\newblock Changes not for the fainthearted: reorienting health care systems toward health equity through action on the social determinants of health.
\newblock American journal of public health. 2009;99(11):1967-74.

\bibitem{patra2021extracting}
Patra BG, Sharma MM, Vekaria V, Adekkanattu P, Patterson OV, Glicksberg B, et~al.
\newblock Extracting social determinants of health from electronic health records using natural language processing: a systematic review.
\newblock Journal of the American Medical Informatics Association. 2021;28(12):2716-27.

\bibitem{ong2024artificial}
Ong JCL, Seng BJJ, Law JZF, Low LL, Kwa ALH, Giacomini KM, et~al.
\newblock Artificial intelligence, ChatGPT, and other large language models for social determinants of health: Current state and future directions.
\newblock Cell Reports Medicine. 2024;5(1).

\bibitem{winden2018evaluation}
Winden TJ, Chen ES, Monsen KA, Wang Y, Melton GB.
\newblock Evaluation of flowsheet documentation in the electronic health record for residence, living situation, and living conditions.
\newblock AMIA Summits on Translational Science Proceedings. 2018;2018:236.

\bibitem{Hollister2017developmen}
Hollister B, Restrepo N, Farber-Eger E, Crawford D, Aldrich M, Non A.
\newblock Development and performance of text-mining algorithms to extract socioeconomic status from deidentified electronic health records.
\newblock In: Pacific Symposium on Biocomputing; 2017. .

\bibitem{Wu2023natural}
Wu W, Holkeboer KJ, Kolawole TO, Carbone L, Mahmoudi E.
\newblock Natural language processing to identify social determinants of health in Alzheimer's disease and related dementia from electronic health records.
\newblock Health Services Research. 2023;58(6):1292-302.

\bibitem{amrit2017identifying}
Amrit C, Paauw T, Aly R, Lavric M.
\newblock Identifying child abuse through text mining and machine learning.
\newblock Expert systems with applications. 2017;88:402-18.

\bibitem{feller2020detecting}
Feller DJ, Don't~Walk OJB, Zucker J, Yin MT, Gordon P, Elhadad N, et~al.
\newblock Detecting social and behavioral determinants of health with structured and free-text clinical data.
\newblock Applied clinical informatics. 2020;11(01):172-81.

\bibitem{lybarger2021annotating}
Lybarger K, Ostendorf M, Yetisgen M.
\newblock Annotating social determinants of health using active learning, and characterizing determinants using neural event extraction.
\newblock Journal of Biomedical Informatics. 2021;113:103631.

\bibitem{HAN2022103984}
Han S, Zhang RF, Shi L, Richie R, Liu H, Tseng A, et~al.
\newblock Classifying social determinants of health from unstructured electronic health records using deep learning-based natural language processing.
\newblock Journal of Biomedical Informatics. 2022;127:103984.

\bibitem{guevara2024large}
Guevara M, Chen S, Thomas S, Chaunzwa TL, Franco I, Kann BH, et~al.
\newblock Large language models to identify social determinants of health in electronic health records.
\newblock NPJ digital medicine. 2024;7(1):6.

\bibitem{tan2024large}
Tan Z, Beigi A, Wang S, Guo R, Bhattacharjee A, Jiang B, et~al.
\newblock Large language models for data annotation: A survey.
\newblock arXiv preprint arXiv:240213446. 2024.

\bibitem{adler2015patients}
Adler NE, Stead WW.
\newblock Patients in context—EHR capture of social and behavioral determinants of health.
\newblock Obstetrical \& Gynecological Survey. 2015;70(6):388-90.

\bibitem{bettencourt2020discovering}
Bettencourt-Silva JH, Mulligan N, Sbodio M, Segrave-Daly J, Williams R, Lopez V, et~al.
\newblock Discovering new social determinants of health concepts from unstructured data: framework and evaluation.
\newblock In: Digital Personalized Health and Medicine. IOS Press; 2020. p. 173-7.

\bibitem{park2021discovering}
Park Y, Mulligan N, Gleize M, Kristiansen M, Bettencourt-Silva JH.
\newblock Discovering associations between social determinants and health outcomes: merging knowledge graphs from literature and electronic health data.
\newblock In: AMIA Annual Symposium Proceedings. vol. 2021. American Medical Informatics Association; 2021. p. 940.

\bibitem{bettencourt2020exploring}
Bettencourt-Silva JH, Mulligan N, Jochim C, Yadav N, Sedlazek W, Lopez V, et~al.
\newblock Exploring the social drivers of health during a pandemic: Leveraging knowledge graphs and population trends in COVID-19.
\newblock In: Integrated Citizen Centered Digital Health and Social Care. IOS Press; 2020. p. 6-11.

\bibitem{gleize2021social}
Gleize M, Mulligan N, Di~Bari A, Bettencourt-Silva JH.
\newblock Social determinant trends of covid-19: An analysis using knowledge graphs from published evidence and online trends.
\newblock In: Public Health and Informatics. IOS Press; 2021. p. 744-8.

\bibitem{chandak2023building}
Chandak P, Huang K, Zitnik M.
\newblock Building a knowledge graph to enable precision medicine.
\newblock Scientific Data. 2023;10(1):67.

\bibitem{li2024dalkdynamiccoaugmentationllms}
Li D, Yang S, Tan Z, Baik JY, Yun S, Lee J, et~al.. DALK: Dynamic Co-Augmentation of LLMs and KG to answer Alzheimer's Disease Questions with Scientific Literature; 2024.

\bibitem{li2024contextualization}
Li D, Tan Z, Chen T, Liu H.
\newblock Contextualization Distillation from Large Language Model for Knowledge Graph Completion.
\newblock In: Findings of the Association for Computational Linguistics: EACL 2024; 2024. p. 458-77.

\bibitem{GPT-4o}
OpenAI. Hello GPT-4o; 2024.
\newblock Available from: \url{https://openai.com/index/hello-gpt-4o/}.

\bibitem{pu2023graph}
Pu Y, Beck D, Verspoor K.
\newblock Graph embedding-based link prediction for literature-based discovery in Alzheimer’s Disease.
\newblock Journal of Biomedical Informatics. 2023;145:104464.

\bibitem{li2022graph}
Li MM, Huang K, Zitnik M.
\newblock Graph representation learning in biomedicine and healthcare.
\newblock Nature Biomedical Engineering. 2022;6(12):1353-69.

\bibitem{Huang2023.03.19.23287458}
Huang K, Chandak P, Wang Q, Havaldar S, Vaid A, Leskovec J, et~al.
\newblock A foundation model for clinician-centered drug repurposing.
\newblock medRxiv. 2024.

\bibitem{kipf2016semi}
Kipf TN, Welling M.
\newblock Semi-supervised classification with graph convolutional networks.
\newblock arXiv preprint arXiv:160902907. 2016.

\bibitem{wei2024pubtator}
Wei CH, Allot A, Lai PT, Leaman R, Tian S, Luo L, et~al.
\newblock PubTator 3.0: an AI-powered literature resource for unlocking biomedical knowledge.
\newblock Nucleic Acids Research. 2024:gkae235.

\bibitem{neumann2019scispacy}
Neumann M, King D, Beltagy I, Ammar W.
\newblock ScispaCy: fast and robust models for biomedical natural language processing.
\newblock arXiv preprint arXiv:190207669. 2019.

\bibitem{healthypeople}
2030 HP. Social Determinants of Health; 2024.
\newblock Available from: \url{https://health.gov/healthypeople/objectives-and-data/social-determinants-health}.

\bibitem{theobald2009extraction}
Theobald M, Shah N, Shrager J.
\newblock Extraction of conditional probabilities of the relationships between drugs, diseases, and genes from PubMed guided by relationships in PharmGKB.
\newblock Summit on Translational Bioinformatics. 2009;2009:124.

\bibitem{bordes2013translating}
Bordes A, Usunier N, Garcia-Duran A, Weston J, Yakhnenko O.
\newblock Translating embeddings for modeling multi-relational data.
\newblock Advances in neural information processing systems. 2013;26.

\bibitem{trouillon2016complex}
Trouillon T, Welbl J, Riedel S, Gaussier {\'E}, Bouchard G.
\newblock Complex embeddings for simple link prediction.
\newblock In: International conference on machine learning. PMLR; 2016. p. 2071-80.

\bibitem{sun2019rotate}
Sun Z, Deng ZH, Nie JY, Tang J.
\newblock Rotate: Knowledge graph embedding by relational rotation in complex space.
\newblock arXiv preprint arXiv:190210197. 2019.

\bibitem{szklarczyk2023string}
Szklarczyk D, Kirsch R, Koutrouli M, Nastou K, Mehryary F, Hachilif R, et~al.
\newblock The STRING database in 2023: protein--protein association networks and functional enrichment analyses for any sequenced genome of interest.
\newblock Nucleic acids research. 2023;51(D1):D638-46.

\bibitem{tran2020heterogeneous}
Tran VD, Sperduti A, Backofen R, Costa F.
\newblock Heterogeneous networks integration for disease--gene prioritization with node kernels.
\newblock Bioinformatics. 2020;36(9):2649-56.

\bibitem{rao2018phenotype}
Rao A, Vg S, Joseph T, Kotte S, Sivadasan N, Srinivasan R.
\newblock Phenotype-driven gene prioritization for rare diseases using graph convolution on heterogeneous networks.
\newblock BMC medical genomics. 2018;11:1-12.

\bibitem{valentini2014extensive}
Valentini G, Paccanaro A, Caniza H, Romero AE, Re M.
\newblock An extensive analysis of disease-gene associations using network integration and fast kernel-based gene prioritization methods.
\newblock Artificial Intelligence in Medicine. 2014;61(2):63-78.

\bibitem{renaux2023knowledge}
Renaux A, Terwagne C, Cochez M, Tiddi I, Now{\'e} A, Lenaerts T.
\newblock A knowledge graph approach to predict and interpret disease-causing gene interactions.
\newblock BMC bioinformatics. 2023;24(1):324.

\bibitem{notterman2015epigenetics}
Notterman DA, Mitchell C.
\newblock Epigenetics and understanding the impact of social determinants of health.
\newblock Pediatric Clinics. 2015;62(5):1227-40.

\bibitem{adkins2023structural}
Adkins-Jackson PB, George KM, Besser LM, Hyun J, Lamar M, Hill-Jarrett TG, et~al.
\newblock The structural and social determinants of Alzheimer's disease related dementias.
\newblock Alzheimer's \& Dementia. 2023;19(7):3171-85.

\bibitem{udayar2022lysosomal}
Udayar V, Chen Y, Sidransky E, Jagasia R.
\newblock Lysosomal dysfunction in neurodegeneration: emerging concepts and methods.
\newblock Trends in neurosciences. 2022;45(3):184-99.

\end{thebibliography}
\end{document}